\documentclass[sigplan,screen]{acmart}

\usepackage{amsmath,epsfig}
\usepackage{algorithm,amsfonts}
\usepackage{subfigure}
\usepackage{graphicx}
\usepackage{booktabs}
\usepackage{bbding}
\usepackage{array}
\usepackage{multirow}

\begin{document}

\copyrightyear{2021}
\acmYear{2021}
\setcopyright{acmcopyright}\acmConference[MMAsia '20]{ACM Multimedia Asia}{March 7--9, 2021}{Virtual Event, Singapore}
\acmBooktitle{ACM Multimedia Asia (MMAsia '20), March 7--9, 2021, Virtual Event, Singapore}
\acmPrice{15.00}
\acmDOI{10.1145/3444685.3446320}
\acmISBN{978-1-4503-8308-0/21/03}

\title{Change Detection from SAR Images Based on Deformable Residual Convolutional Neural Networks}

\author{Junjie Wang}
\affiliation{%
 \institution{Ocean University of China}}
\email{wangjunjie4945@stu.ouc.edu.cn}

\author{Feng Gao}
\authornote{Corresponding author: Feng Gao.}
\affiliation{%
 \institution{Ocean University of China}}
\email{gaofeng@ouc.edu.cn}

\author{Junyu Dong}
\affiliation{%
 \institution{Ocean University of China}}
\email{dongjunyu@ouc.edu.cn}

\begin{abstract}

Convolutional neural networks (CNN) have made great progress for synthetic aperture radar (SAR) images change detection. However, sampling locations of traditional convolutional kernels are fixed and cannot be changed according to the actual structure of the SAR images. Besides, objects may appear with different sizes in natural scenes, which requires the network to have stronger multi-scale representation ability. In this paper, a novel \underline{D}eformable \underline{R}esidual Convolutional Neural \underline{N}etwork (DRNet) is designed for SAR images change detection. First, the proposed DRNet introduces the deformable convolutional sampling locations, and the shape of convolutional kernel can be adaptively adjusted according to the actual structure of ground objects. To create the deformable sampling locations, 2-D offsets are calculated for each pixel according to the spatial information of the input images. Then the sampling location of pixels can adaptively reflect the spatial structure of the input images. Moreover, we proposed a novel pooling module replacing the vanilla pooling to utilize multi-scale information effectively, by constructing hierarchical residual-like connections within one pooling layer, which improve the multi-scale representation ability at a granular level. Experimental results on three real SAR datasets demonstrate the effectiveness of the proposed DRNet.

\end{abstract}



\maketitle

\section{Introduction}

Remote sensing image change detection aims to accurately detect the changed information of the same geographical area by analyzing two images captured at different times. It plays an essential role in many applications, such as target detection \cite{Bai10}, disaster monitoring \cite{Martino07}, and natural resource supervision \cite{Bruzzone97}.

Benefiting from the rapid development of the earth observation programs, more and more multi-temporal SAR images are available. These SAR images are produced by an active system that sends a signal to the ground and receives the reflected signal. Since the active sensors are capable of observing the earth all-weather and all-time, SAR images have become the ideal data source for change detection.

In the past few years, many SAR images change detection methods are proposed. These methods can be broadly divided into two directions: supervised and unsupervised \cite{Chen19}. The supervised methods generally perform better since more prior information is taken into account. However, supervised methods require a large volume of high-quality labeled samples, and sample collection is time-consuming. Therefore, it is nearly impossible to perform supervised change detection in practice. The unsupervised methods are more preferable. Therefore, we mainly focus on unsupervised change detection in this paper.

Existing unsupervised change detection methods are generally comprised of three steps: 1) preprocessing, 2) difference image (DI) generation, 3) DI classification. In the first step, geometric correction and registration are usually involved and play an important role. In the second step, the ratio-based method \cite{Bazi05} is commonly used to generate a DI. Many efficient operators are proposed, such as the log-ratio operator \cite{Dekker98}, Gauss-ratio operator \cite{Hou14}, and neighbor-based-ratio operator \cite{Gong12}. In the step of DI classification, many methods based on convolutional neural networks (CNNs) have been proposed. Liu \emph{et al.} \cite{Liu18} presented a change detection framework based on convolutional coupling network. Mou \emph{et al.} \cite{Mou19} proposed a change detection framework which combines CNNs and recurrent neural network into one end-to-end network. Gao \emph{et al.} \cite{Gao19} detected changed information from sea ice SAR images by transferred deep CNNs. Liu \emph{et al.} \cite{Liu19} presented a local restricted CNN framework for SAR change detection, in which the original CNNs are improved with the local spatial constraint.

However, there are some shortcomings in these methods. On the one hand, these methods use a fixed grid in the sampling location of the convolutional kernels, and the spatial features of the complex structure are not fully exploited. On the other hand, these methods do not fully learn the information of objects with different sizes, and the multi-scale representation ability by utilizing features with different resolutions is insufficient.

\begin{figure*}[ht]
\begin{center}
\includegraphics [width=7in]{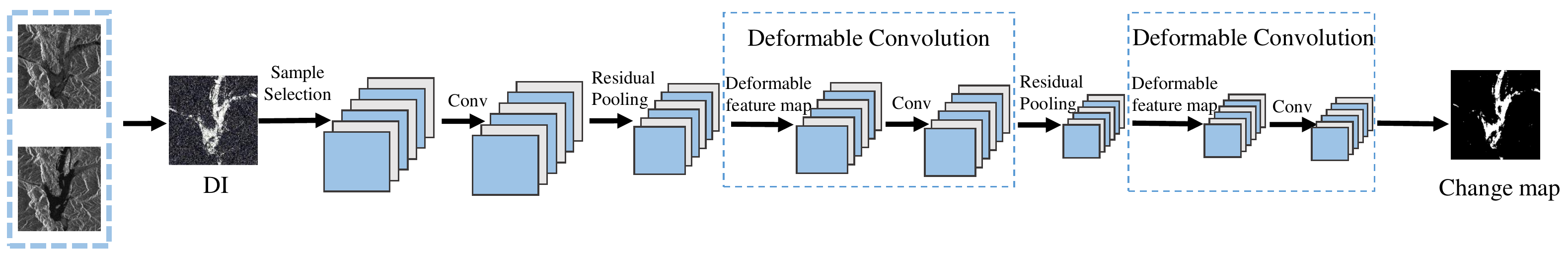}
\caption{Illustration of the architecture of the DRNet}
\label{structure}
\end{center}
\end{figure*}

\begin{figure}
\begin{center}
\includegraphics [width=2.5in]{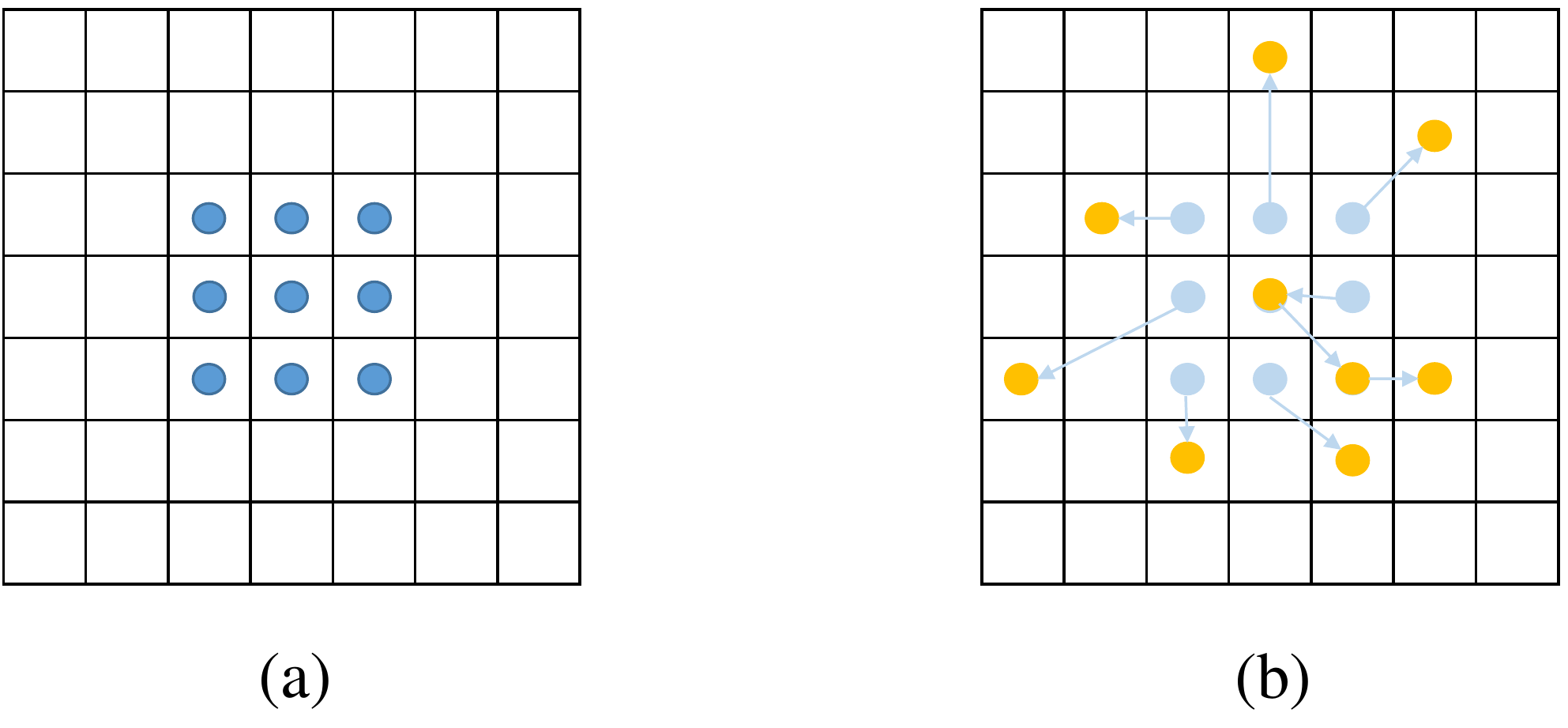}
\caption{Illustration of the sampling locations in $3\times3$ standard and deformable convolutions. (a) Regular sampling grid of standard convolution. (b) Deformable sampling locations.}
\label{deformable}
\end{center}
\end{figure}

To tackle the aforementioned limitations of existing CNN-based change detection methods, we present a \underline{D}eformable \underline{R}esidual Convolutional Neural \underline{N}etwork (DRNet) to detect the changed information from multitemporal SAR images. On the one hand, we introduce deformable convolutional sampling locations \cite{Dai17} to extract features of different shapes of ground objects. The deformable convolutional sampling locations can be obtained by calculating the 2-D offsets of each pixel of the input image. As illustrated in Fig. \ref{deformable}. The offsets are learned from the preceding feature maps via additional convolutional layers. Thus, the deformation is conditioned on the input features in a local, dense, and adaptive manner. On the other hand, we design a novel pooling module named residual pooling, which extracts multi-scale features on a granular level. We replace the pooling kernel of $n$ channels with a set of kernel group, each with $w$ channels ($n=s\times w$). Then these features after pooling are connected in a hierarchical residual-like style to increase the number of scales that output features can represent. Extensive experiments are conducted on three SAR datasets to validate the effectiveness of the proposed DRNet. The results demonstrate that the proposed DRNet can well adapt to the complex spatial structure of SAR images and achieves better performance compared with the state-of-the-art methods.

In summary, the main contributions of the proposed DRNet are as follows:

\begin{itemize}

\item We introduce deformable convolution into SAR change detection which can adaptively change the sampling location according to the complex spatial structures, and thus improve the performance of change detection.

\item To explore the multi-scale information of the ground objects, we design a novel pooling module-residual pooling, which is simple and flexible variants of pooling for boosting the multi-scale representation ability of CNNs. It utilizes a set of smaller pooling kernels with $w$ channels to replace the vanilla pooling with $n$ channels, then it connects these kernels in a hierarchical residual-like style, which increases the number of scales that output features can represent.

\item We conducted extensive comparative experiments on three SAR datasets to validate the effectiveness of DRNet. The results demonstrate that the proposed DRNet can represent multi-scale information better and adaptively adjust the convolution shape according to the structure of objects.

\end{itemize}

\section{Methodology}

In this paper, we devote to address the problem of change detection from multitemporal SAR images. Given two coregistered SAR images $I_1$ and $I_2$ captured in the same geographical area at different times, we aim to generate a change map which shows the change information between two images. It is generally treated as a binary classification task. In the generated change map, the change pixels are marked as ``1", and unchanged pixels are marked as ``0".

The proposed change detection method has two advantages compared to traditional CNN-based methods. First, we introduce the deformable convolution to capture the complex spatial information of SAR images. Besides, we design a novel pooling module - residual pooling to fuse multi-scale information in the pooling layer. In this section, we first describe the implementation of deformable convolution, then present the detailed structure of residual pooling.

\subsection{Deformable Convolution}

The CNN-based SAR change detection methods employ a stack of layers to extract features from the input images and identify the change information \cite{Cheng17}. However, the convolutional kernels of regular CNNs can hardly adapt to the spatial structure of input images, since the sampling locations of the convolutions are fixed grids ($N  \times N$ pixels).

\begin{figure}
\begin{center}
\includegraphics [width=3.4in]{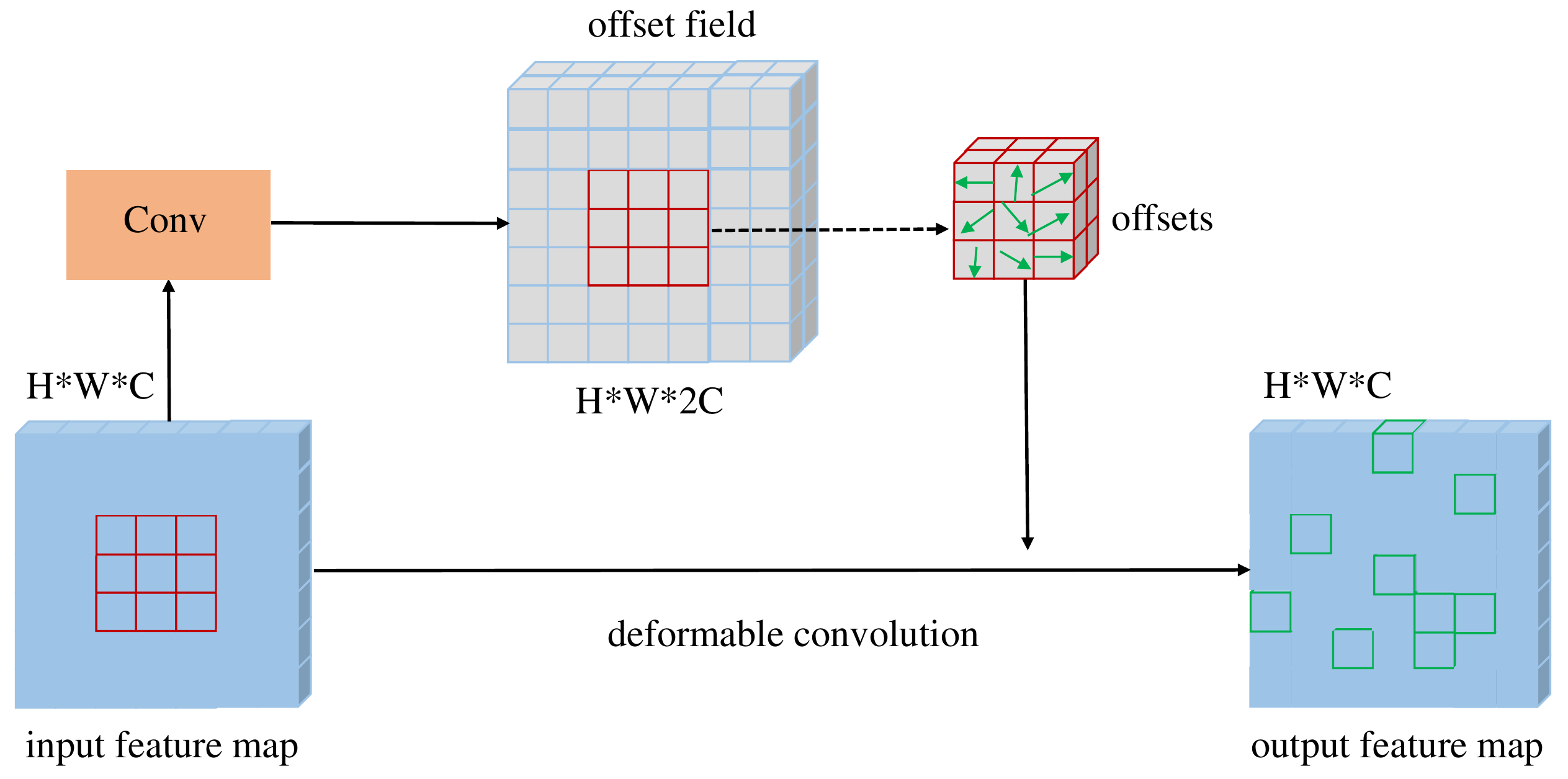}
\caption{Illustration of  $3\times3$  deformable convolutions.}
\label{deformableconvolution}
\end{center}
\end{figure}

To address this issue, we introduce the deformable convolution \cite{Dai17} to adjust the sampling location. In the network, we add two steps before regular convolution: offset field generation and deformable feature map generation. As is shown in Fig. \ref{deformableconvolution}, the offsets are learned from the preceding feature maps, via additional convolutional layers. Given a pixel in an input feature map with location $(x,y)$ and value $p(x,y)$, it corresponds to two values $\Delta x$ and $\Delta y$ in the offset fields. A deformable feature map fuses the neighboring similar pixels. The new pixel value can be denoted by:

\begin{equation}
\begin{split}
 p_\textrm{new}(x,y)=p(\min(\max(0, x+\Delta x), N-1),~ \\
 \min(\max(0, y +\Delta y), N-1)).
\end{split}
\end{equation}

The weight of the convolutional filters for generating offset fields are trained based on spatial features to enable the sampling locations to be transferred to the neighboring similar pixels. Finally, regular convolutions are employed on the deformable feature maps, and the output can be obtained as follows:
\begin{equation}
{y(x,y)}={\sum_{s_w}w_{ij}}\cdot p_{new}(x_{ij},y_{ij}),
\end{equation}
where $w_{i,j}$ represents the corresponding weight of the kernel and $s_w$ enumerates all locations of the kernel.

\subsection{Residual pooling}

Multi-scale feature representation is of great importance for numerous vision tasks. Recent advances in backbone convolutional neural networks (CNNs) continually demonstrate that stronger multi-scale representation ability results in better visual understanding. However, the common solution of multi-scale feature extraction includes making CNNs deeper or building multiple branches, which increase the complexity and the amount of parameter of model, requiring more time and resources to optimize.

To avoid these inconveniences, Huang \emph{et al.}\cite{Huang18} proposed stacked pooling which focuses on the pooling module to extract multi-scale information. It can capture the multi-scale responses by stacking multiple pooling kernels. The advantage of using pooling layer to extract multi-scale features is non-parametric and it makes the network as simple as possible. However, in \cite{Huang18}, the features are extracted in a layer-wise manner. As mentioned in \cite{Xie}, in addition to the dimension of width and depth, the cardinality of features is also a concrete, measurable dimension that is of central importance. Increasing cardinality is a more effective way of gaining accuracy than making network deeper or wider. Inspired by this idea and Gao's work \cite{Gao19_arxiv}, we proposed residual pooling, improving the multi-scale representation capability at a more granular level, which is orthogonal to exiting methods that utilize layer-wise operations. Thus, the proposed residual pooling can be easily plugged into most exiting CNN models replacing the vanilla pooling.

\begin{figure}
\begin{center}
\includegraphics [width=3.3in]{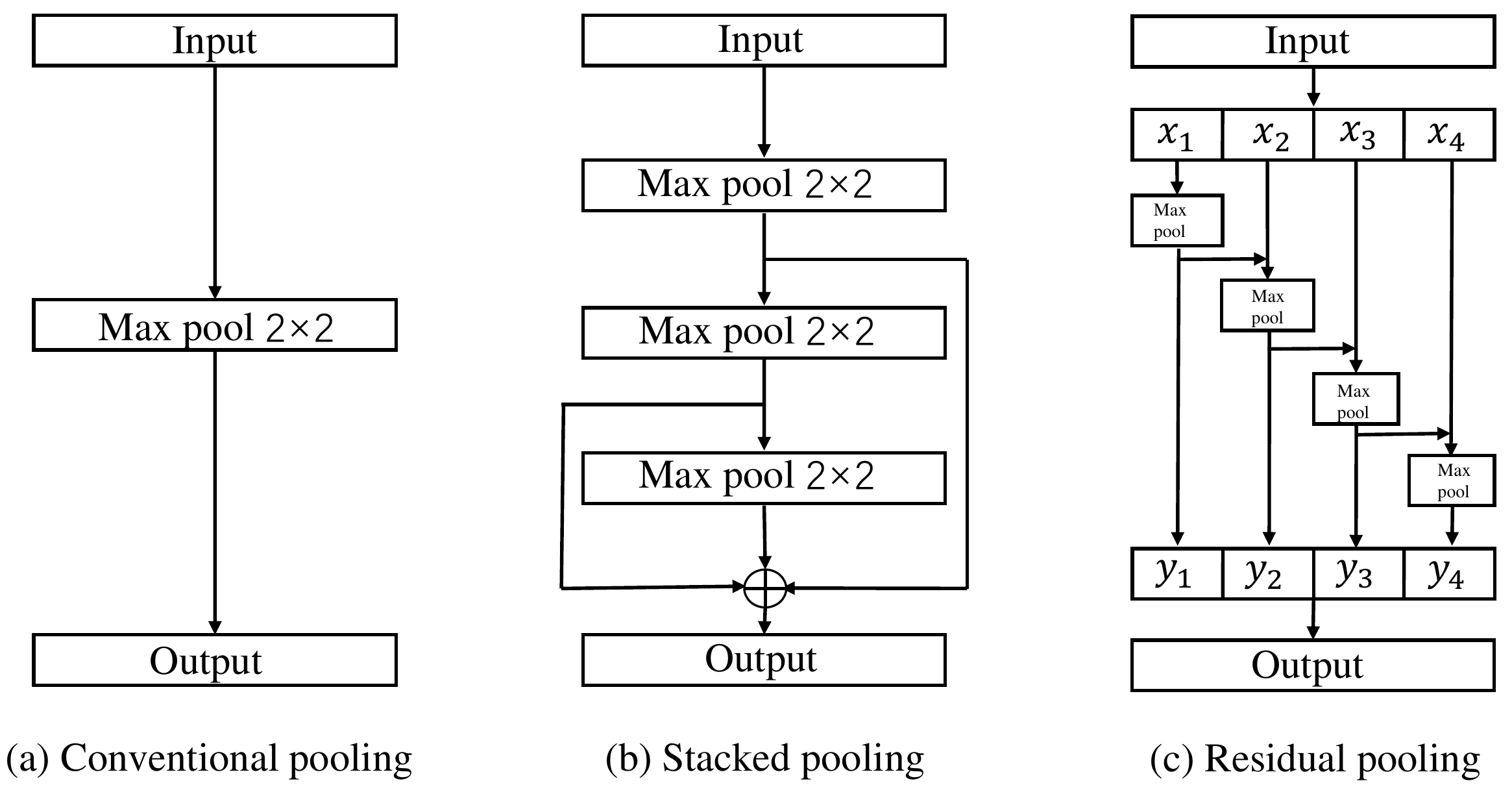}
\caption{Comparison among the conventional pooling, stacked pooling, and residual pooling. The 2 $\times$ 2 max-pooling operator is taken as an example.}
\label{pooling}
\end{center}
\end{figure}

To achieve this goal, we replace the max-pooling operator of $n$ channels with $s$ feature map subsets, and each subset contain $w$ channels ($n = s \times w$). As is shown in Fig. \ref{pooling}, these smaller pooling subsets are connected in a hierarchical residual-like style to increase the number of scales that the output features can represent. Specifically, we evenly divide input feature maps into $s$ feature map subsets. denoted by $x_i$, where $i \in (1,2,...,s)$. Each feature subset $x_i$ has the same spatial size but $1/s$ number of channels compared with the input feature map. Each $x_i$ has a corresponding $2\times2$ max-pooling operator, denoted by $K_i(\cdot)$. $y_i$ is used to denote the output of $K_i(\cdot)$. The feature subset $x_i$ is added with the output of $K_{i-1}(\cdot)$ and then fed into $K_i(\cdot)$. Therefore, $y_i$ can be defined by:
\begin{equation}
y_i=\left\{
\begin{array}{rcl}
K_i(x_i)         &  &  {i=1;} \\
K_i(x_i+y_{i-1}) &  &  {2\leq i \leq s;}\\
\end{array}
\right.
\end{equation}

\section{Experimental Results and Analysis}

In this section, we first describe the datasets and the evaluation criteria used in our experiments. Then, some investigation of several important parameters on the change detection performance is presented. After that, we make some ablation experiments to verify the effect of deformable convolution and residual pooling. Finally, the proposed method is compared with several state-of-the-art methods.

\subsection{Dataset Description and Evaluation Criteria}

\begin{figure}
\begin{center}
\includegraphics [width=3.3in]{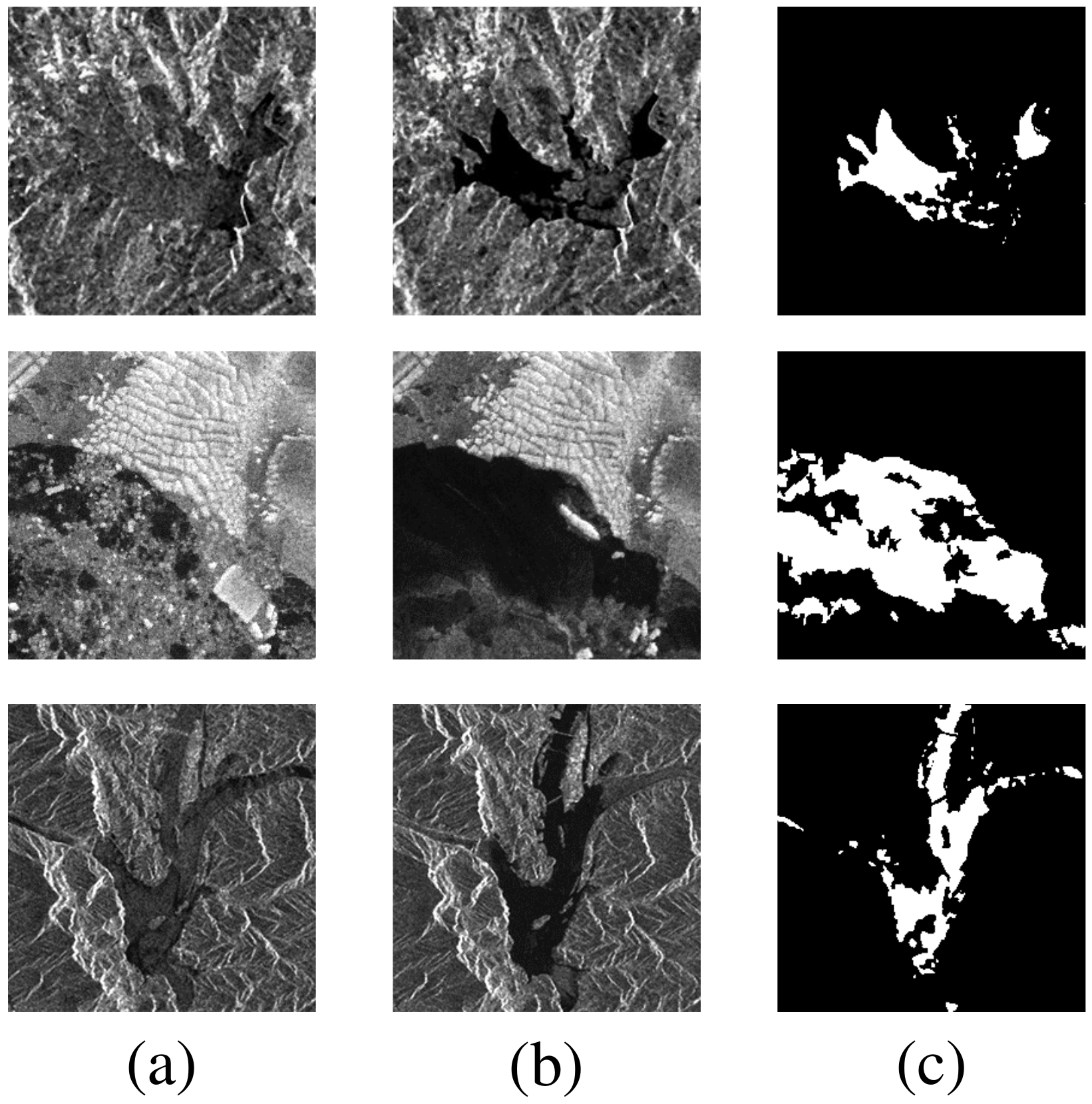}
\caption{The datasets used in our experiment.}
\label{dataset}
\end{center}
\end{figure}

In Fig. \ref{dataset}, we show the three datasets used in our experiments. Figures (a) and (b) represent SAR images captured at different times, and (c) represents the ground truth image.   The first line of Fig. \ref{dataset} is Florence dataset, which is captured over the city of Florence, Italy. It illustrate an area with a varied and complex morphology. The second line represents the the Sulzberger dataset. It show the process of the sea ice breakup. The last dataset is the Seoul dataset, which is selected from two SAR images captured over Seoul, South Korea.

The performance evaluation of change detection is a critical issue. We use false positives (FP), false negatives (FN), overall error (OE), percentage of correct classification (PCC) as indicators to measure the effectiveness of the proposed method. The FP is the number of pixels that are unchanged class in the ground truth image but wrongly classified as changed ones. The FN is the number of pixels that are changed class in the ground truth image but wrongly classified as unchanged ones. The OE is computed by OE $=$ FP $+$ FN. The PCC is computed by $\textrm{PCC} = (1 -\textrm{OE}/Nt)\times 100\%$, where $Nt$ represents the total pixels in the ground truth image.

\subsection{Analysis of the crucial parameters of DRNet}

\subsubsection{Analysis of different number of training samples}

The number of training samples is an essential parameter in the SAR image change detection, since most deep learning-based methods need a large number of samples to optimize parameters. In this subsection, we show the relationship between PCC and training samples number on three datasets.

\begin{figure}[ht]
\centering
\includegraphics [width=3.2in]{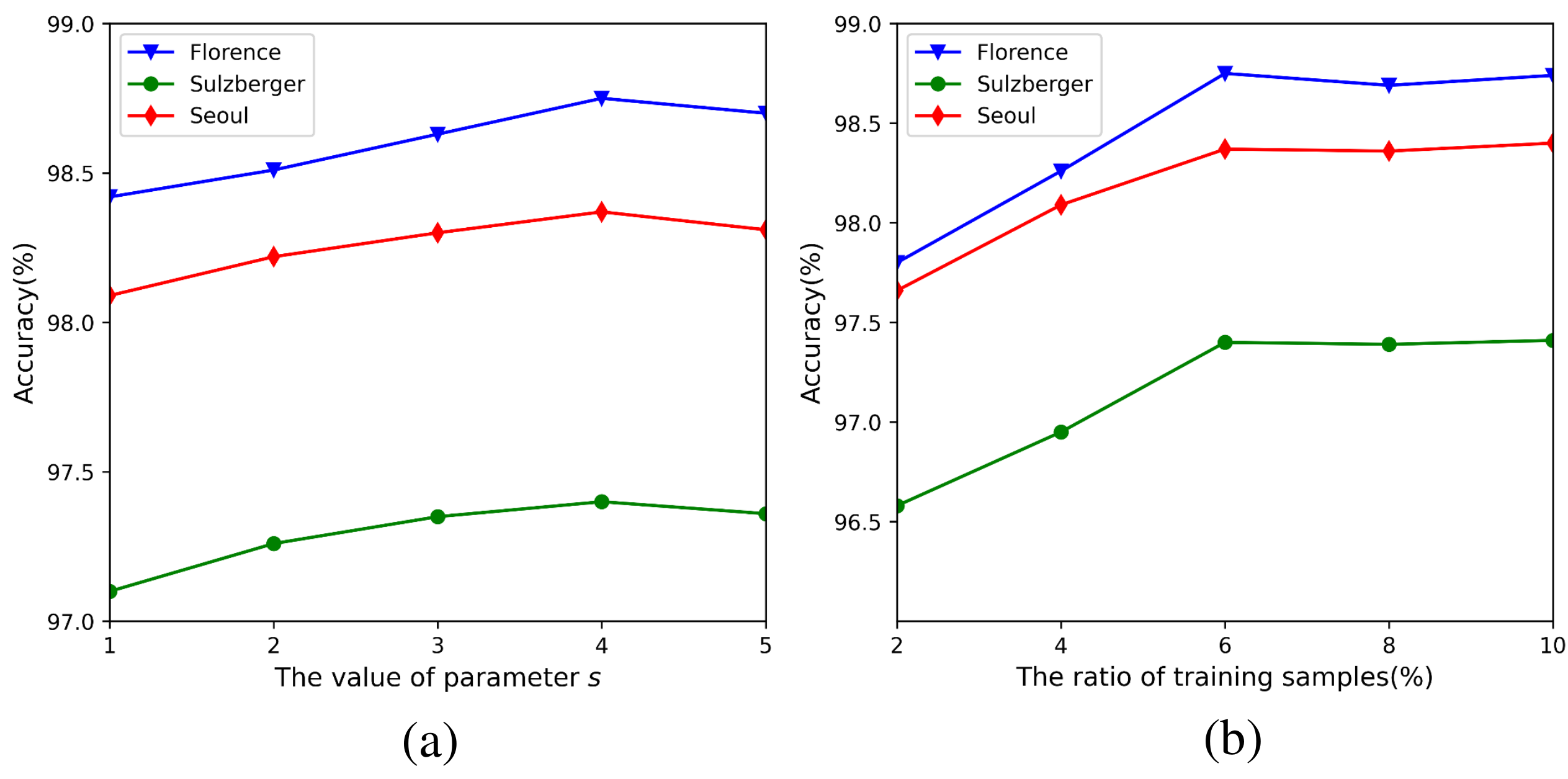}
\caption{Relationship between PCC and the crucial parameters.}
\label{parameter}
\end{figure}

To verify the validity of the model, we randomly selected 2\%, 4\%, 6\%, 8\%, and 10\% pixels as training samples, respectively. Fig. \ref{parameter} (a) shows the quantitative analysis result of the relationship between PCC and the number of training samples. From these curves, we can see that there is a sharp increase in the PCC value when the number of training samples from 2\% to 6\%. After that, the PCC value tends to become stable even the number of training samples becomes larger. Thus, we select 6\% pixels as training samples. Compared with other methods \cite{Gao19_rs}\cite{Gao19_grsl}, DRNet needs fewer samples because the structure of the model is relatively simple, and the proposed residual pooling does not introduce additional parameters.

\subsubsection{Analysis of the parameter $s$}

The parameter $s$ denotes the number of subsets after the feature maps are sent to the residual pooling, and it is a critical parameter in residual pooling. We evaluate the classification performance by taking $s$=1, 2, 3, 4, and 5, respectively. The PCC value is investigated as the validation criterion on three datasets.

From Fig. \ref{parameter} (b), we can observe that when $s=4$, the proposed DRNet obtains the best results. It indicates that the multi-scale information can be extracted effectively by dividing the feature maps into groups for pooling of different scales, leading to the improvement of performance. It should be noted that when $s=1$, the residual pooling is equal to the vanilla pooling. By comparing the results of $s=1$ and $s=4$, we can confirm that the residual pooling can replace vanilla pooling flexibly, and obtain some improvement in the field of change detection.

\subsection{Ablation Studies}

We further discuss and validate the effectiveness of the deformable convolution and residual pooling on three datasets. It should be noted that in order to make a fair comparison, we have designed a $Basic\ network$ with the same structure as DRNet, which uses regular convolution and vanilla pooling replacing deformable convolution and residual pooling. The results are shown in Table \ref{ablation}.

\begin{table*}[htbp]
\centering
\caption{Ablation experiments on three datasets.}
\renewcommand\arraystretch{1.5}
\begin{tabular}{cccccccc}
\toprule
\# & Basic network & \shortstack{Deformable\\convolution} & Stacked pooling & Residual pooling & Florence dataset & Sulzberger dataset & Seoul dataset \\ \midrule
1  & \Checkmark & \XSolidBrush & \XSolidBrush & \XSolidBrush & 98.26 & 96.95 & 97.84 \\
2  & \Checkmark & \Checkmark & \XSolidBrush & \XSolidBrush & 98.42  & 97.10 & 98.09 \\
3  & \Checkmark & \Checkmark & \Checkmark & \XSolidBrush & 98.56  & 97.23 & 98.18 \\
4  & \Checkmark & \Checkmark & \XSolidBrush & \Checkmark & 98.75 & 97.40 & 98.37 \\
5  & \Checkmark & \XSolidBrush & \Checkmark & \XSolidBrush & 98.33 & 97.05 & 97.91 \\
6  & \Checkmark & \XSolidBrush & \XSolidBrush & \Checkmark & 98.43 & 97.13 & 98.04 \\
\bottomrule
\end{tabular}
\label{ablation}
\end{table*}

As reported in Table \ref{ablation}, although the $Basic\ network$ is a deep learning-based model, due to the fixed sampling locations of traditional convolution and the limitation of the receptive field, the performance of $Basic\ network$ is not satisfying. When the deformable convolution is introduced into the network, it can adaptively adjust the shape of convolution, reflecting the complex structure more effectively, resulting in the improvement of performance (\#1 vs \#2). Moreover, the advances in backbone CNN caused by stronger multi-scale representation ability demonstrated a trend toward more effective and efficient multi-scale representations. There, we proposed residual pooling, which does not introduce extra parameters and can easily take place of the vanilla pooling layer. In order to verify its effectiveness, we compare it with vanilla pooling (\#1 vs \#6). Through the experiment, we can know that perceiving information from different scales is essential which further indicates the advantage of residual pooling compared with vanilla pooling. It also should be noted that compared with residual pooling, when we introduce stacked pooling (\#5), the experiment results also are improved, but not as good as residual pooling. The reason for this phenomenon is that feature maps contain different granularity features, and unified pooling will lead to the loss of information. After grouping feature maps, residual pooling can not only expand the receptive field, but also better explore the relevance of information between different feature maps. The proposed DRNet which combines the deformable convolution and residual pooling has received the best result on three datasets, not only showing their respective effectiveness, but also showing the complementary effects between them (\#4).

\subsection{Results on the Three Datasets}

In order to verify the effectiveness of the proposed DRNet, we compare it with six closely related methods: PCAKM \cite{Celik09}, NBRELM\cite{Gao16}, GaborPCANet \cite{GaoGRSL16}, MLFN\cite{Gao19_grsl}, LR-CNN \cite{Liu19}, and DBN \cite{Gong16}. In PCAKM, the contextual information is taken into account by principal component analysis (PCA), and the extracted features are clustered by the k-means algorithm. NBRELM utilizes NR operator to obtain some pixels belonging to changed or unchanged class, then these pixels are put into ELM to train a model. GaborPCANet is a simplified deep learning model which is comprised of two PCA convolutional layer, binary hashing layer and block-wise histogram generation layer. In MLFN, a large dataset is used to train a model, and the deep knowledge can be transferred to change detection. LR-CNN is a convolutional neural network with local spatial restrictions on the output layer. In DBN, a deep belief network is employed to complete the SAR change detection task, which produces a change detection map directly from two images with a trained deep neural network.
In order to more intuitively and adequately reflect the results of various methods, both visual and quantitative analyses are made in our experiments. For visual analysis, the change maps generated by different methods are exhibited in figure form. For quantitative analysis, the change maps are exhibited in tabular form.

\begin{figure*}[ht]
\begin{center}
\includegraphics [width=7in]{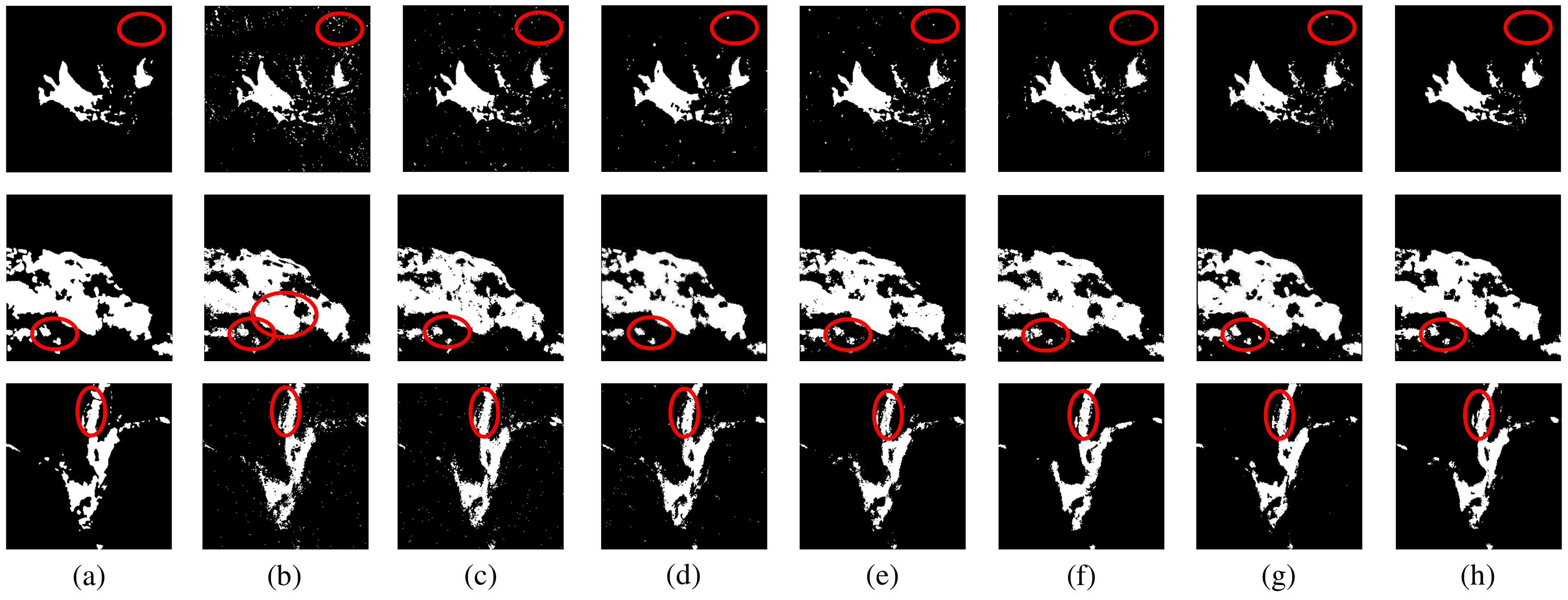}
\caption{Visualized results of different change detection methods on three datasets. (a) Ground truth image. (b) Result by PCAKM \cite{Celik09}. (c) Result by NBRELM \cite{Gao16}. (d) Result by GaborPCANet \cite{GaoGRSL16}. (e) Result by MLFN \cite{Gao19_grsl}. (f) Result by LR-CNN \cite{Liu19}. (g) Result by DBN \cite{Gong16}. (h) Result by the proposed DRNet.}
\label{result}
\end{center}
\end{figure*}

\begin{table*}[t]
\centering
\caption{Change detection results of different methods on the three datasets.}
\label{table1}
\begin{tabular}{|c|c|c|c|c|c|c|c|c|c|c|c|c|c|}
\hline
\multicolumn{2}{|c|}{\multirow{2}{*}{Method}} &
\multicolumn{4}{c|}{Florence}&\multicolumn{4}{c|}{Sulzberger}&\multicolumn{4}{c|}{Seoul}\\
\cline{3-14}

\multicolumn{1}{|c}{} && FP & FN & OE & PCC(\%) & FP & FN & OE & PCC(\%) & FP & FN & OE & PCC(\%) \\
\hline

\multicolumn{2}{|c|}{PCAKM \cite{Celik09}} & 1307 & 663 & 1970 & 96.99	     & 2368 & 203 & 2571 & 96.08      & 1013 & 998 & 2011 & 96.93\\ 
\hline
\multicolumn{2}{|c|}{NBRELM \cite{Gao16}} & 1186 & 598 & 1784 & 97.28	    & 1860 & 579 & 2439 & 96.28      & 871 & 963 & 1834 & 97.20\\ 
\hline
\multicolumn{2}{|c|}{GaborPCANet \cite{GaoGRSL16}}& 1062 & 579 & 1642 & 97.50	& 1410 & 783 & 2193 & 96.65    & 547 & 926 & 1473 & 97.75\\ 
\hline
\multicolumn{2}{|c|}{MLFN \cite{Gao19_grsl}} & 851 & 406 & 1257 & 98.08 	& 1254 & 819 & 2073 & 96.84         & 492 & 875 & 1367 & 97.91\\ 
\hline
\multicolumn{2}{|c|}{LR-CNN \cite{Liu19}} & 793 & 364 & 1157 & 98.23       & 1198 & 680 & 1878 & 97.13             & 420 & 854 & 1274 & 98.06\\ 
\hline
\multicolumn{2}{|c|}{DBN \cite{Gong16}} & 725 & 330 & 1055 & 98.39  	& 632 & 1242 & 1874 & 97.14            & 362 & 869 & 1231 & 98.12\\ 
\hline
\multicolumn{2}{|c|}{Proposed DRNet} & 603 & 215 & 818 & 98.75 & 621 & 1183 & 1704 & 97.40                  & 343 & 725 & 1068 & 98.37\\ 
\hline

\end{tabular}
\label{table_result}
\end{table*}

Fig. \ref{result} illustrates the final change maps of different methods on the three datasets, the first row shows the results of different methods on Florence dataset, and the next two rows are Sulzberger and Seoul datasets. Table \ref{table_result} lists the corresponding evaluation criteria of different methods. From the visual comparison, it can be observed that just like the area marked by the red circle, there are some noisy white spots in the results generated by PCAKM, NBRELM, GaborPCANet, MLFN and LR-CNN, which corresponds to the higher FP value in the Table \ref{table_result}. What's more, it can be seen that the PCC and OE value of deep learning-based methods (MLFN, LR-CNN and DBN) is generally better. It means that deep learning can enhance model comprehension better, thereby reducing detection errors. Besides, from Table \ref{table_result}, we can observe that the proposed DRNet has the minimum FP and FN value, leading to that the proposed DRNet can provide the most similar results to the ground-truth change map compared with other methods. On three datasets, the OE and PCC values of DRNet performs better than other methods, the PCC values have increased by at least 0.36\%, 0.26\% and 0.25\%, respectively. It further proves that the proposed DRNet can adapt to the complex structure of the dataset and improve the multi-scale representation ability.

\section{Conclusion}

In this paper, a deformable residual convolutional neural network (DRNet) is proposed for the task of SAR image change detection. Compared to existing CNN-based methods, the DRNet makes improvements in two aspects. First, we introduce deformable convolution to extract more effective spatial features according to the complex spatial structure of the ground objects. Besides, we design multi-scale residual pooling to utilize multi-scale information at a granular level. Compared with other methods using multi-scale information, we focus on the pooling layer, which not only not introduces additional parameters, but also can easily take place of the vanilla pooling layer in the implementation. Compared to vanilla pooling, the residual pooling can not only expand the receptive field, but also capture different granularity features. Experimental results on three real SAR datasets demonstrate that the proposed DRNet has better performance than some state-of-the-art methods in terms of both the visual quality of the change map and evaluation criteria.

\section*{Acknowledgement}

The authors would like to thank the anonymous reviewers for their helpful and insightful comments. This work is supported by
the National Key Research and Development Program of China (No. 2018AAA0100602), the National Natural Science Foundation of China (No. U1706218), and the Key Research and Development Program of Shandong Province (No. 2019GHY112048).


\end{document}